\documentclass[letterpaper, 10pt, conference]{ieeeconf}      

\IEEEoverridecommandlockouts                              
                                                          
\usepackage{graphicx} 

\usepackage[hidelinks]{hyperref}

\makeatletter
\newcommand\notsotiny{\@setfontsize\notsotiny\@vipt\@viipt}
\makeatother

\title{An Earth Rover dataset recorded at the ICRA@40 party}
\author{\authorblockN{Qi Zhang $^{\href{https://orcid.org/0000-0003-4064-695X}{\includegraphics[scale=0.05]{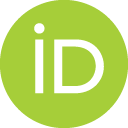}}}$} \authorblockA{\textit{University of Amsterdam} \\ \textit{The Netherlands}} \and \authorblockN{Zhihao Lin 
$^{\href{https://orcid.org//0009-0005-8152-3886}{\includegraphics[scale=0.05]{Figures/ORCIDiD_icon128x128.png}}}$}
\authorblockA{\textit{University of Glasgow} \\ \textit{United Kingdom}} \and \authorblockN{Arnoud Visser $^{\href{https://orcid.org/0000-0002-7525-7017}{\includegraphics[scale=0.05]{Figures/ORCIDiD_icon128x128.png}}}$} \authorblockA{\textit{University of Amsterdam} \\ \textit{The Netherlands}} }
\date{June 2024}

\begin{document}

\maketitle

\begin{abstract}
    The ICRA conference is celebrating its $40^{th}$ anniversary in Rotterdam in September 2024, with as highlight the Happy Birthday ICRA Party at the iconic Holland America Line Cruise Terminal. One month later the IROS conference will take place, which will include the Earth Rover Challenge. In this challenge open-world autonomous navigation models are studied truly open-world settings.

    As part of the Earth Rover Challenge several real-world navigation sets in several cities world-wide, like Auckland, Australia and Wuhan, China.
    The only dataset recorded in the Netherlands is the small village Oudewater. The proposal is to record a dataset with the robot used in the Earth Rover Challenge in Rotterdam, in front of the Holland America Line Cruise Terminal, before the festivities of the Happy Birthday ICRA Party start. 

    \textbf{See:} {\scriptsize \url{https://github.com/SlamMate/vSLAM-on-FrodoBots-2K} }
\end{abstract}

\section{Introduction}

Recent developments in visual SLAM have shown that it is possible to generate detailed reconstructions of the surroundings (e.g. \cite{liso2024loopy} and \cite{yugay2024gaussianslam}), although those reconstructions require depth-cameras. Creating navigation maps from a single front camera is more challenging, with the first successful attempt called MonoSLAM \cite{Davison2007monoslam}, followed by several more advanced methods like LDSO, ORB-SLAM, DROID-SLAM and SNI-SLAM \cite{Gao2018ldso, campos2021orb, teed2021droid, zhu2024sni}. Some more advanced approaches, which integrate multi-view information and use neural networks to replace depth sensors for depth estimation \cite{zhang2024glorieslamgloballyoptimizedrgbonly, sandstrom2024splatslamgloballyoptimizedrgbonly}, applying more constraints with additional features such as lines or planes \cite{7989522, 10238802}, or extracting more general features \cite{s23042113, BRUNO202197}, have shown promising prospects for real-time monocular SLAM methods.

\begin{figure}[b]
        \centering
                \includegraphics[height=0.15\textheight]{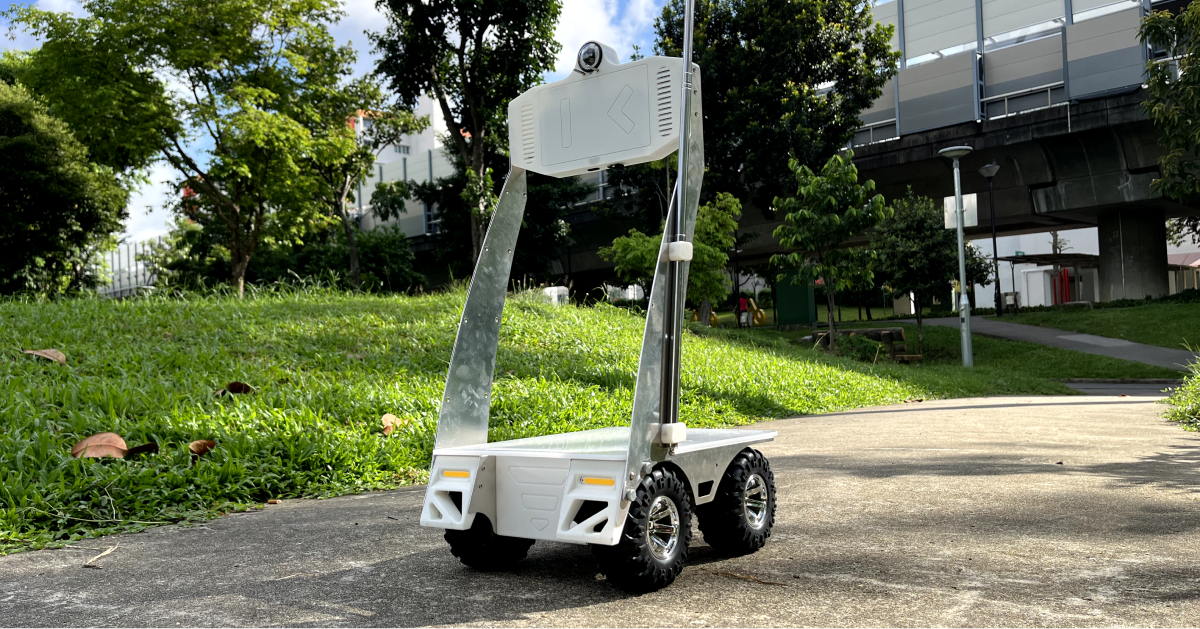}
\caption{The FrodoBot Earth Rover Zero.}
        \label{fig:frodobot}
\end{figure}

The robot used for this challenge is the Earth Rover Zero from FrodoBot
(see Fig.~\ref{fig:frodobot}). It is a small four-wheel drive robot, with a camera at the front and rear. The front camera has a resolution of resolution of 1024x576, and the rear camera's resolution is 540x360. In addition, the FrodoBot has an IMU, GPS and 4G SIM card onboard. The 4G SIM card provides a wireless interface, which allows teams all over the world to drive the robot around.

Part of the challenge is creating a real-world dataset ($>2k$ hours), collected in several cities. Such large datasets are the way forward,
as demonstrated with foundations models for robotics \cite{padalkar2023open}.

Such a dataset could be used in two ways. One solution would be to use it to train the robot to navigate with an end-to-end algorithm \cite{9309347, 9981574}, as demonstrated recently in the OpenDriveLab challenge by the team of Nvidia \cite{li2024hydramdpendtoendmultimodalplanning}.
Yet, in that challenge, six high-resolution cameras provided a surround view, while the Earth Rover has to rely mostly on its front camera.

\begin{figure}[tb]
        \centering
                \includegraphics[height=0.166\textheight]{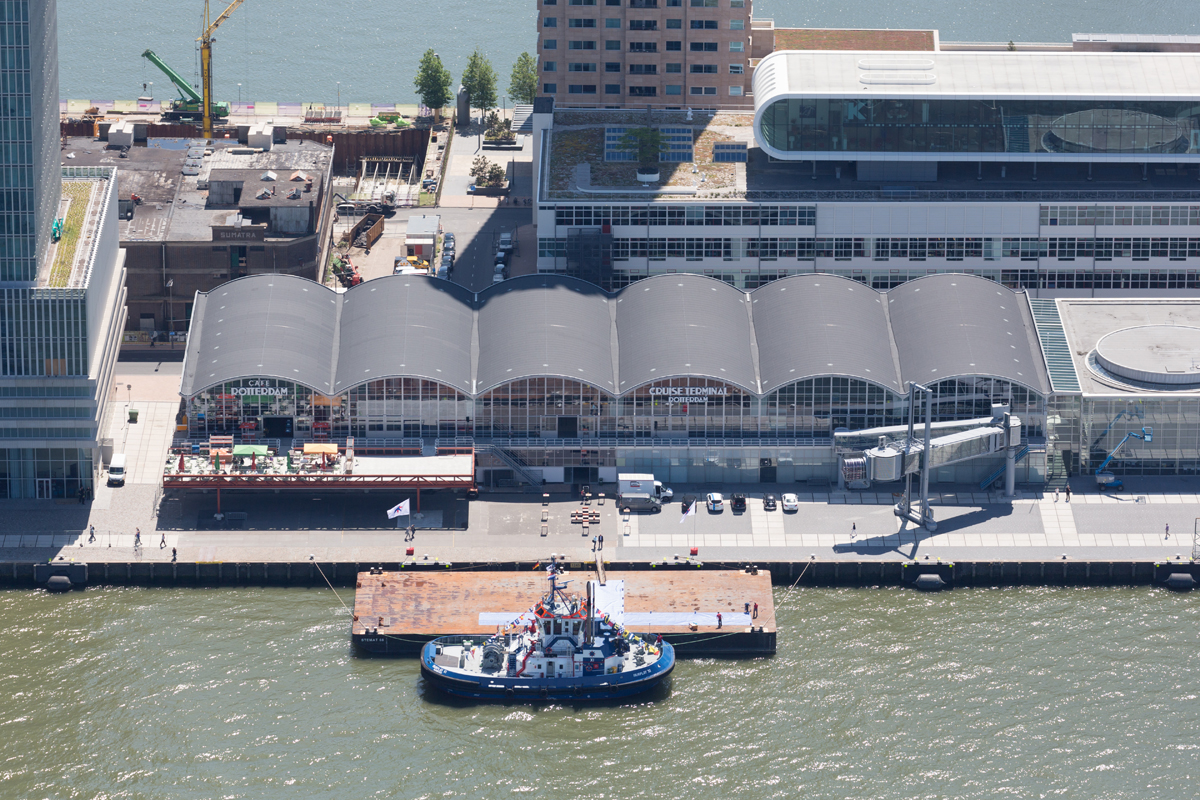}
\caption{Holland America Line
Cruise Terminal in Rotterdam.}
        \label{fig:cruiseterminal}
\end{figure}

\begin{figure*}[tb]
        \centering
                \vspace*{0.8em}
                \includegraphics[width=1.95\columnwidth]{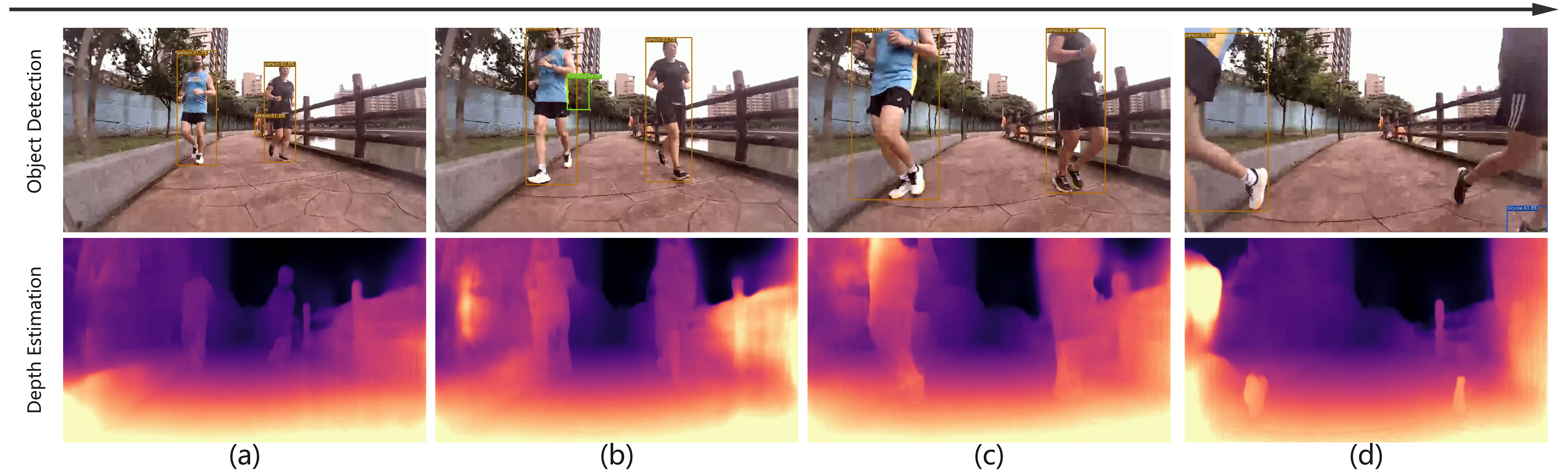} 
\caption{The first row shows the target boxes obtained by YOLOX from the frames. The second row of images shows the dense depth maps predicted by LiteMono from a single RGB image.}
        \label{fig:yolox}
\end{figure*}

The front camera can also be used for situation awareness, by perceiving the features that provide information for safe navigation. For safety, it is for instance important that pedestrians, bicycles and other traffic are detected \cite{Cao2022pedestrian}. For a first demonstration, a robust object-detection model such as YoloV8 would be a good choice \cite{kumar2023object}. Rich semantic priors can eliminate the interference of dynamic objects in pose estimation \cite{10477312, 9981826}. In contrast, richer semantic methods, instance segmentation, multi-object tracking or more complex networks, further combine the estimation of the poses of dynamic objects and the robot itself \cite{Bescs2020DynaSLAMIT, 9982280, Ye_2023_CVPR}. In addition, the front camera can be used for visual odometry, to get an estimate of the Earth Rover's movement, independent from the IMU estimates \cite{kneip2011robust}. We can also integrate IMU and visual information to achieve more robust tracking and reconstruction \cite{AZIMI202218, 9981795}.

\section{Dataset}

At the day of the ICRA@40 party we recorded two datasets in Rotterdam. One was a track along the street called 'Wilhelminakade', between the Cruise Terminal and the Metro station. The other track was waiting on the guests before the entrance of the Cruise Terminal, entering the building after the guest arrived, and cruising between the guests to the dancefloor. One week later we recorded a tour around the venue, including a view over the harbor from the 'Holland-Amerikakade'. This part of Rotterdam is a small peninsula, with for vehicles no option for through traffic (a one way street that circles back). Bicycles and pedestrians have the option to pass through over a small bridge at the other end of the peninsula, so together with hotels, bars and restaurants it is a lively city district. The robot stayed mostly at the sidewalk, although it also crossed the 'Wilhelminakade'. There was some construction work going on, so the sidewalk was not always 4m wide. Part of the sidewalk (along the Cruise Terminal) is a gallery, so a GPS signal was not guaranteed.

The three separate dataset recordings can be found at the following location:

\begin{itemize}
    \item \href{https://doi.org/10.21942/uva.27130089.v3}{before the ICRA@40 party - 223 MB}
    \item \href{https://doi.org/10.21942/uva.27127125.v2}{during the ICRA@40 party - 308 MB}
    \item \href{https://doi.org/10.21942/uva.27159765.v5}{after the ICRA@40 party - 7.68 GB}
\end{itemize}

The three datasets combined can be downloaded from \href{https://dx.doi.org/10.21227/0r3g-cs83}{IEEE Dataport}.

All sensor readings, including the front and back camera, are recorded via the SDK (version 4.3.13)\footnote{\scriptsize \url{https://github.com/frodobots-org/earth-rovers-sdk} }, which means that our dataset was recorded at the laptop controlling the Earth Rover. Previous recordings\footnote{\scriptsize \url{https://huggingface.co/datasets/frodobots/FrodoBots-2K} \label{note:dataset}} were  directly recorded at the Earth Rover itself. Our method reflects more closely the conditions of the Earth Rover Challenge. This also means that the image frames are received with a maximum frame-rate of 3Hz. 

The first two datasets and the last dataset are not post-processed with the same method. The last dataset is larger than the previous recordings, because we used an optical flow method \cite{Huang2022} to interpolate between frames to get a higher FPS result (6 Hz).  In the last recording only the faces are blurred, based on face detection with the YuNet algorithm \cite{Wu2023}. This is in contrast with the first two recordings, where the whole body of the person is detected based on Yolov8M\footnote{\notsotiny \url{https://github.com/ultralytics/ultralytics/releases/tag/v8.1.0}} and blurred. In all datasets depth-images are included, although the  Earth Rover is not equipped with a depth camera. This depth is estimated with the LiteMono \cite{Zhang_2023_CVPR}, as described in the next section.

\section{Situation Awareness}

The next level would be to create a visual map of the surroundings, in this case, the Holland America Line
Cruise Terminal (see Fig.~\ref{fig:cruiseterminal}). The ORB-SLAM algorithm \cite{campos2021orb} would be a good choice for a first demonstration\footnote{
\scriptsize \url{https://github.com/SlamMate/vSLAM-on-FrodoBots-2K} \label{note:github}}, which can be extended in a later stage with more advanced methods \cite{zhu2024sni}.

Once a visual map is generated, it can be combined with satellite maps and street maps to Scene Action Maps \cite{loo2024scene}.
The behavior learned to navigate on the Scene Action Maps, can be ported to other robots with cross-embodiment approaches \cite{yang2024crossembodiment, doi:10.1126/scirobotics.abg5810}.

We have many interesting ideas how to extend the autonomous driving capabilities of the robot, but also realize that well ahead of the Earth Rover Challenge only the basic navigational and safety functionality will be active. For instance, dynamic-obstacle avoidance is hard; demonstrating this in a real-world setting will give insight in the strength and weaknesses.  With human supervision this will be enough to record an interesting dataset \cite{bu2024fieldnotesdeployingresearch}.

The initial experiments$^{\ref{note:github}}$
the largest models in YOLOX \cite{yolox2021} and LiteMono \cite{Zhang_2023_CVPR}, YOLOX-X, and LiteMono8m, are used to demonstrate the performance of the largest real-time models on the dataset. Additionally, we use the state-of-the-art visual SLAM system ORBSLAM3 to process the information in the dataset. In the dataset, We selected the output\_rides\_21 sequence from the FrodoBots-2K dataset$^{\ref{note:dataset}}$ for our experiments because this sequence is highly similar to the port environment, featuring sparse moving objects and repetitive textures near the port. Additionally, this sequence was recorded along the riverside using the same robot.

The top row of Fig.~\ref{fig:yolox} shows the initial results of object detection with YOLOX-X. In the first three frames the two persons are correctly identified. Only in the last frame there are two False Negatives, when the two persons are to close by to be recognized. In addition False Positives can be seen in this frames. In the first frame from left to right, the flagpole is mistakenly identified as a person. In the second frame, the tree is mistakenly identified as a handbag.

In the bottom row of Fig.~\ref{fig:yolox} shows the initial results of the depth estimation with LiteMono8m. The ground-plane can be seen, and the fence at the right is clearly visible. The two persons approaching can also been seen, although the module shows some blurriness in the depth boundaries of the person.

\begin{figure}[hb]
        \centering
                \includegraphics[width=\columnwidth]{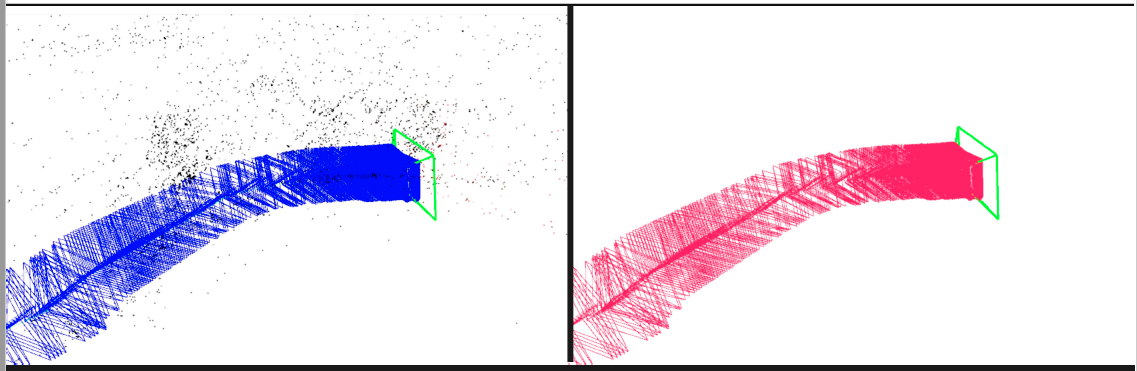}
\caption{A part of the trajectory plot of ORBSLAM3 during operation on the output\_rides\_21 sequence (see Fig.~\ref{fig:yolox}). Squares represent individual frames, with red indicating tracking failure frames and blue indicating tracking success frames.}
        \label{fig:orbslam3}
\end{figure}

In Fig.~\ref{fig:orbslam3} the initial results of the ORBSLAM3 algorithm are presented, Blue represents good tracking, while red indicates tracking failure. It can be seen that ORBSLAM3 frequently experiences tracking failures throughout the entire dataset because of the repetitive textures along the river channel and drift trajectories due to dynamic objects in the scene. This demonstrates the significant challenges this dataset poses to existing real-time technologies and prompts us to reconsider the capabilities of current techniques.

\section{Conclusion}

This paper outlines plans to record a new dataset for the Earth Rover Challenge at ICRA@40 in Rotterdam using the Earth Rover Zero robot. The dataset will enrich the current collection\footnote{\scriptsize \url{https://huggingface.co/datasets/frodobots/FrodoBots-2K}} further which will allow to train SLAM and navigation algorithms on data collected in the real world. This recording will be unique due to the public (the visitors of the Happy Birthday ICRA Party) and the waterfront nearby. As a bonus, it will be fun for the ICRA participants to see a live dataset recording in action during their party.

\pagebreak

\IEEEtriggeratref{29}
\bibliographystyle{IEEEtran}
\bibliography{IEEEabrv,references}

\end{document}